\documentclass[10pt,twocolumn,letterpaper]{article}

\usepackage[pagenumbers]{cvpr} 

\usepackage{graphicx}
\usepackage{amsmath}
\usepackage{amssymb}
\usepackage{booktabs}
\usepackage{times}
\usepackage{epsfig}
\usepackage{algorithm}
\usepackage{algpseudocode}
\usepackage{listings}
\usepackage{lineno}
\usepackage{color}
\usepackage{paralist}
\usepackage{array}
\usepackage{enumitem}
\usepackage{multirow}

\usepackage[pagebackref, breaklinks, colorlinks, bookmarks=false, citecolor=blue,linkcolor=blue, urlcolor=blue]{hyperref}

\definecolor{darkgreen}{RGB}{30,150,30}
\definecolor{darkblue}{RGB}{0,0,127}
\definecolor{darkyellow}{RGB}{171,133,0}
\definecolor{darkred}{RGB}{180,20,20}
\definecolor{darkmagenta}{RGB}{127,0,127}
\definecolor{darkcyan}{RGB}{0,127,127}

\def\red #1{\textcolor[rgb]{1,0,0}{\textbf{#1}}}
\def\blue #1{\textcolor[rgb]{0,0,1}{\textbf{#1}}}
\def\green #1{\textcolor[rgb]{0,0.5,0}{\textbf{#1}}}

\usepackage[capitalize]{cleveref}
\crefname{section}{Sec.}{Secs.}
\Crefname{section}{Section}{Sections}
\Crefname{table}{Table}{Tables}
\crefname{table}{Tab.}{Tabs.}

\def\cvprPaperID{1558} 
\def\confName{CVPR}
\def\confYear{2022}

\begin{document}

\title{ELSA: Enhanced Local Self-Attention for Vision Transformer}

\author{
Jingkai Zhou$^{12}$\thanks{Work done during an internship at Alibaba Group.}~~~Pichao Wang$^{2}$\thanks{Corresponding author, project lead}~~~Fan Wang$^{2}$~~~Qiong Liu$^{1}$\thanks{Corresponding author.}~~~Hao Li$^{2}$~~~Rong Jin$^{2}$
\\{$^1$South China University of Technology~~~$^2$Alibaba Group}\\
\small{201510105876@mail.scut.edu.cn, liuqiong@scut.edu.cn}\\
\small\{pichao.wang, fan.w, lihao.lh, jinrong.jr\}@alibaba-inc.com
}
\maketitle

\begin{abstract}
Self-attention is powerful in modeling long-range dependencies, but it is weak in local finer-level feature learning. The performance of local self-attention (LSA) is just on par with convolution and inferior to dynamic filters, which puzzles researchers on whether to use LSA or its counterparts, which one is better, and what makes LSA mediocre.
To clarify these, we comprehensively investigate LSA and its counterparts from two sides: \emph{channel setting} and \emph{spatial processing}. We find that the devil lies in the generation and application of spatial attention, where relative position embeddings and the neighboring filter application are key factors.
Based on these findings, we propose the enhanced local self-attention (ELSA) with Hadamard attention and the ghost head. Hadamard attention introduces the Hadamard product to efficiently generate attention in the neighboring case, while maintaining the high-order mapping. The ghost head combines attention maps with static matrices to increase channel capacity.
Experiments demonstrate the effectiveness of ELSA. Without architecture / hyperparameter modification, drop-in replacing LSA with ELSA boosts Swin Transformer~\cite{swin} by up to +1.4 on top-1 accuracy. 
ELSA also consistently benefits VOLO~\cite{volo} from D1 to D5, where ELSA-VOLO-D5 achieves 87.2 on the ImageNet-1K without extra training images.
In addition, we evaluate ELSA in downstream tasks. ELSA significantly improves the baseline by up to +1.9 box Ap / +1.3 mask Ap on the COCO, and by up to +1.9 mIoU on the ADE20K. Code is available at \url{https://github.com/damo-cv/ELSA}.
\end{abstract}

\vspace{-3mm}
\section{Introduction}
\label{sec:intro}

From upstream to downstream visual tasks, vision transformers~\cite{vit, deit, detr, setr, transtrack, trans_reid} have set off a revolution by achieving promising results. Behind the success, the multi-head self-attention (MHSA) plays a critical role, which generates attention maps to dynamically aggregate spatial information, leading to greater flexibility and larger capacity. Recent studies ~\cite{do_vit_see_like_cnn, t2t} 
demonstrated that MHSA tends to focus on local information in the first few layers of vision transformers. Several methods~\cite{coatnet, visformer, volo, vil} introduce inductive bias to force earlier layers to embed local details, which boosts the generalization ability of vision transformers. As a representative of them, Swin Transformer~\cite{swin} brings locality to MHSA and makes great progress on a wide range of visual tasks.

\begin{figure}[t]
\centering 
\includegraphics[width=.95\linewidth]{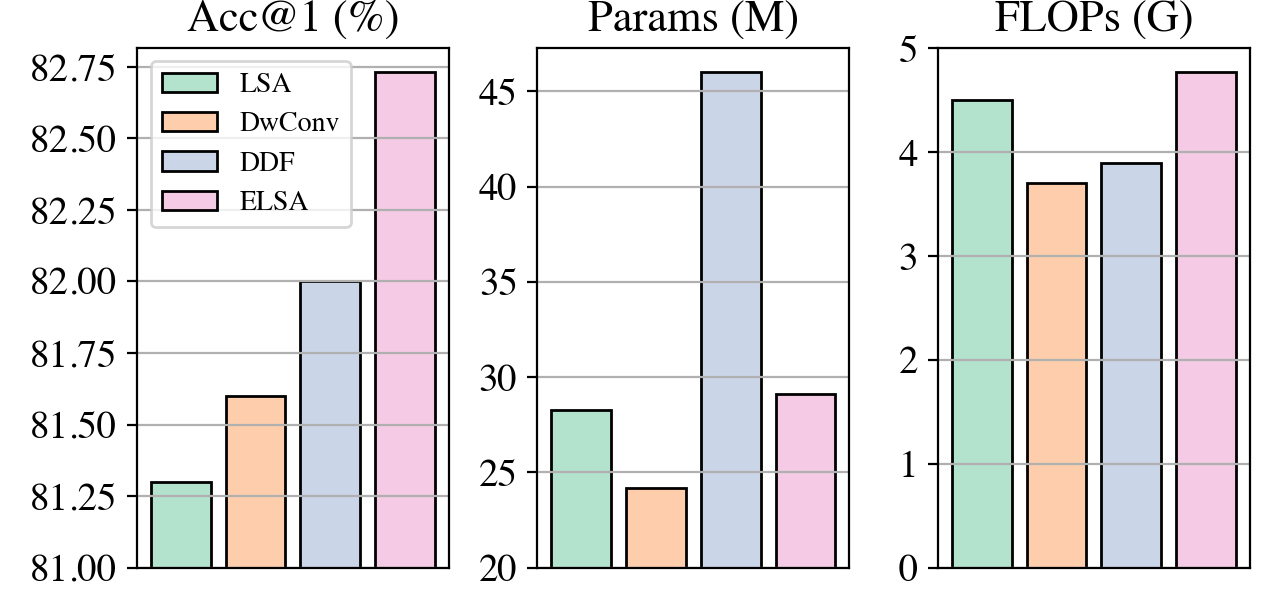}
\caption{\textbf{Comparison of different layers on the Swin-T~\cite{swin} architecture.} Local self-attention (LSA) in Swin-T is less efficient than depth-wise convolution (DwConv). Its top-1 accuracy is also inferior to dynamic filters like DDF~\cite{ddf}. Our ELSA surpasses these counterparts with a large margin while using the similar number of parameters and FLOPs as LSA.} 
\label{fig:local_comp}
\vspace{-3mm}
\end{figure}

However, one strange phenomenon appears in Swin Transformer. One can achieve similar performance when replacing local self-attention (LSA) in Swin Transformer with depth-wise convolution (DwConv)~\cite{xception, mobile_v1} or dynamic filters.
As shown in Figure~\ref{fig:local_comp}, we replace the LSA in Swin-T~\cite{swin} with DwConv and the decoupled dynamic filter (DDF)~\cite{ddf}. LSA only achieves similar top-1 accuracy as DwConv, which is lower than DDF, but it requires more floating-point operations (FLOPs). This phenomenon has also been observed in recent papers~\cite{demystifying, volo, coatnet, what_makes}, but they lack detailed analysis of the reason behind such performances, and it motivates us to raise a question: \textit{what makes local self-attention mediocre?}

To answer this question, we thoroughly review LSA, DwConv, and dynamic filters from two key aspects: \emph{channel setting} and \emph{spatial processing}. 

\vspace{1mm}
\noindent\textbf{Channel setting.} One straightforward difference between DwConv and LSA is the channel setting. DwConv applies different filters to different channels. LSA adopts the multi-head strategy, which shares filters (a.k.a attention maps) within each group of channels. In this work, we consider the channel setting of DwConv as a special case of the multi-head strategy, where the number of heads is set to the number of channels. One guess is that more heads in DwConv might be a critical factor in why it performs comparably to LSA. However, our experiments show that even if we set the head number of DwConv to be the same as that of LSA, DwConv still achieves similar or better accuracy. We also find that directly increasing the head number of LSA will not improve its performance. A new channel strategy is demanded by LSA to further improve accuracy.

\vspace{1mm}
\noindent\textbf{Spatial processing.} How to obtain and apply filters (or attention maps) to gather spatial information is another difference between DwConv, dynamic filters, and LSA. DwConv shares static filters across all feature pixels in a sliding window way. Dynamic filters~\cite{carafe, involution, volo, dynamic_filter, ddf} employ a bypath network, normally a $1\times 1$ convolution, to generate spatial-specific filters, and apply these filters to the neighboring area of each pixel. LSA generates attention maps, which are also a kind of spatial-specific filters, via the dot product of the query and key matrices. LSA applies these attention maps to local windows. In this work, we unify the above three kinds of spatial processing into one paradigm, and fairly investigate them from parameterization, normalization, and filter application. We find that relative position embedding and neighboring filter application are two key factors that affect performance. However, calculating the dot product of queries and keys in the neighboring case is not computational-friendly. An efficient way of filter generation in the neighboring case is needed to replace the dot product while maintaining performance.

\vspace{1mm}
Based on these findings, we propose the enhanced local self-attention module (ELSA) to better embed local information. ELSA is composed of Hadamard attention and the ghost head module. In Hadamard attention, we replace the the dot product with the Hadamard product which is more computational-friendly in the neighboring case while maintaining the high-order mapping. The ghost head combines attention with static ghost head matrices, which effectively and efficiently improves channel capacity and performance.
We empirically validate the performance of ELSA by drop-in replacing LSA / Outlooker~\cite{volo} in Swin Transformer~\cite{swin} / VOLO~\cite{volo}. Without changing the architecture / hyperparameter of other parts, ELSA considerably improves the performance of Swin Transformer and VOLO, while introducing little overhead. In addition, we also demonstrate the superior performance of ELSA in downstream object detection and semantic segmentation tasks.

In short, we make the following contributions in this work:
\vspace{-2mm}
\begin{itemize}[itemsep=-1mm, leftmargin=*]
    \item Extensively investigate the LSA and its counterparts so as to empirically reveal \emph{what factors make LSA mediocre}.
    \item Propose the enhanced local self-attention (ELSA) to better embed local details by introducing Hadamard attention and the ghost head.
    \item Validate ELSA in both upstream and downstream tasks. The use of ELSA in drop-in replacement boosts baseline methods consistently.
\end{itemize}

\section{Related Work}
\label{sec:related}

\vspace{1mm}
\noindent \textbf{Vision Transformers.}
Transformer~\cite{attn_is_all} is first proposed in the NLP task and achieves dominant performance~\cite{bert, gpt}. Recently, the pioneering work ViT~\cite{vit} successfully applies the pure transformer-based architecture to computer vision, revealing the potential of transformer in handling visual tasks. Lots of follow-up studies are proposed~\cite{han2021transformer, d2021convit, yuan2021incorporating, xu2021co, chen2021crossvit, graham2021levit, li2021localvit, kitaev2020reformer, guo2021cmt, fan2021multiscale, rao2021dynamicvit, yu2021glance, el2021xcit, chen2021psvit, xu2021evo, huang2021shuffle, mobilevit, soft, hrvit, mobileformer,dpt,local2global, cswin, hrformer}.
Many of them analyze the ViT~\cite{rethinking, can, gong2021improve, chu2021we, coatnet, vitc, demystifying, t2t, wang2021kvt, do_vit_see_like_cnn} and improve it via introducing locality to earlier layers~\cite{coatnet, volo, halo, visformer, vil, focal_attn, swin}. In particular, Raghu \etal~\cite{do_vit_see_like_cnn} observe that the first few layers in ViTs focus on local information. Li \etal~\cite{t2t} also demonstrate that the first few layers embed local details. Xiao \etal~\cite{vitc} and Wang \etal~\cite{wang2021scaled} find that introducing inductive bias, like convolution stem, can stabilize the training and improve the peak performance of ViTs. Similarly, Dai \etal~\cite{coatnet} marry convolution with ViTs, improving the model generalization ability. Swin Transformer~\cite{swin}, as a milestone, also leverages local self-attention (LSA) to embed detailed information in high-resolution finer-level features. Despite these successes, several studies~\cite{demystifying, coatnet, volo, what_makes} observe that the performance of LSA is just on par with convolution in both upstream and downstream tasks~\cite{demystifying}. The reasons behind this phenomenon are not clear, and in-depth comparisons under the same conditions are valuable.

\vspace{1mm}
\noindent \textbf{Dynamic filters.} 
Convolution and depth-wise convolution~\cite{xception} has been widely used in CNNs~\cite{resnet, mobile_v1, mobile_v2, mobile_v3, efficientnet}, while their content-agnostic nature limits the model flexibility and capacity. To solve this problem, dynamic filters are proposed one after another. One kind of dynamic filter~\cite{weightnet, condconv, dynet, dynamic_conv} predicts coefficients to combine several expert filters which are then shared across all spatial pixels. Another kind of dynamic filter~\cite{dynamic_filter, drconv, carafe, carafe++, involution, pac, ddf, volo} generates spatial-specific filters. Specifically, the dynamic filter networks~\cite{dynamic_filter} use the separate network branches to predict a complete filter at each pixel. PAC~\cite{pac} uses a fixed Gaussian kernel on adapting features to modify the standard convolution filter at each pixel. DRConv~\cite{drconv} extends CondConv~\cite{condconv} to each pixel. CARAFE~\cite{carafe} and CARAFE++~\cite{carafe++} are the dynamic layers for upsampling and downsampling, where a channel-shared 2D filter is predicted at each pixel. Similarly, Involution~\cite{involution} applies the CARAFE-like structure to feature extraction. VOLO~\cite{volo} introduces the Outlooker to embed local details.
DDF~\cite{ddf} decouples dynamic filters to spatial and channel ones, reducing computational overhead while achieving promising results. In this work, we observe that dynamic filters, like DDF~\cite{ddf}, perform superior to LSA. Based on comparison and discussion, we empirically reveal the factors leading to such a phenomenon, and propose enhanced local self-attention (ELSA) to better embed local details.

\section{Channel Setting}
\label{sec:channel_setting}

To figure out \emph{what makes LSA mediocre}, we first focus on one of the most obvious differences between DwConv and LSA, \ie, the channel setting. DwConv applies different filters to each channel. Differently, several dynamic filters~\cite{carafe, involution, volo} and LSA employ the multi-head strategy, which splits channels into multiple groups and shares the same filter within each group. In this work, we consider the setting of DwConv as a special case of the multi-head strategy, where the number of heads is equal to the number of channels. Therefore, comparing channel settings between LSA and DwConv is essentially investigating the performance of different head numbers.

\begin{figure}[t]
\centering 
\includegraphics[width=\linewidth]{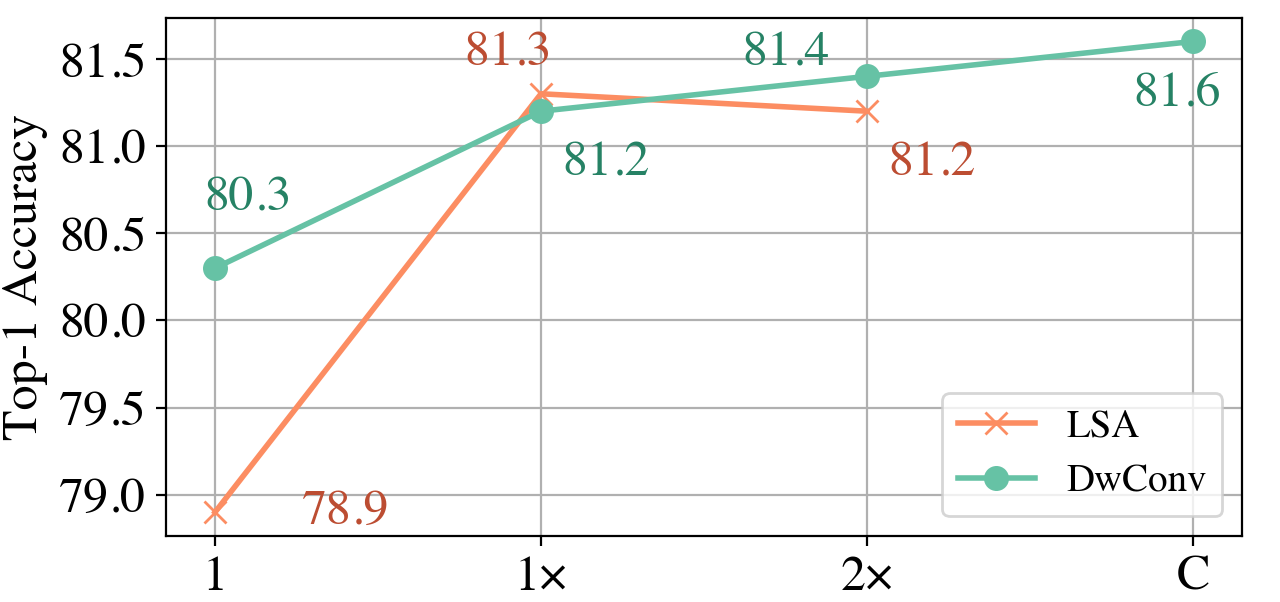}
\caption{\textbf{Investigation of different channel settings.} Only the number of heads is changed each time for fair comparisons. The LSA version runs out of memory under setting C.} 
\label{fig:channel_settings} 
\end{figure}

Figure~\ref{fig:channel_settings} shows the investigation results on the ImageNet-1K dataset. We compare two versions of Swin-T~\cite{swin} under the same hyperparameters, and only modify the head setting each time. Setting 1 means that the number of heads is set to 1 for all layers. Setting 1$\times$ represents the original setting of Swin-T, that is, the head numbers of four stages are set to $\{$3,6,12,24$\}$. Setting 2$\times$ means to double the original setting, \ie, $\{$6,12,24,48$\}$. Setting C denotes that the number of heads is set to the number of channels for all layers.

There has been a guess that more heads in the DwConv version increase channel capacity, thus may lead to comparable performance. However, we find that under the same channel setting, like 1$\times$ and 2$\times$, the DwConv version still achieves similar performance as the LSA one. The DwConv version even exceeds the LSA one under setting 1, which demonstrates that the channel setting is not the essential factor leading to the strange phenomenon. 

In addition, we observe that setting more heads than 1$\times$  does not benefit the LSA version. One possible reason is that more heads lead to fewer channels for each head generation. Thus, directly increasing the number of heads cannot improve channel capacity or performance. A new strategy is desired to further increase channel capacity and accuracy.

\section{Spatial Processing}
As the channel setting is not the critical factor, we seek the answer from the perspective of spatial processing. DwConv, dynamic filters, and LSA adopt different strategies to gather spatial information. We first review these strategies and unify them into one paradigm. Then, we fairly compare these strategies from three aspects.

\subsection{Formulation}

\vspace{1mm}
\noindent\textbf{Conv / DwConv.} Conv and DwConv do not generate filters. They share the static convolutional filters by sliding window. The spatial processing of them can be written as
\begin{equation}
    \textbf{f}_i = \sum_{j \in \Theta} \textbf{w}_{j-i} \textbf{x}_j
    \label{eq:dwconv}
\end{equation}
where $\textbf{f}_i$ represents the output at pixel $i$, $\textbf{x}_j$ represents the input at pixel $j$, $\textbf{w}_{j-i}$ is the filter weight, $j-i$ is the relative offset between pixel $j$ and $i$. Take the filter size 3 as an example, $j-i$ corresponds to $\{(-1,-1),$ $(-1, 0),$ $(-1,1),$ $...,$ $(1, 1)\}$. $\Theta$ notes the neighboring area around the pixel $i$. 

\vspace{1mm}
\noindent\textbf{Dynamic filters.} Dynamic filters~\cite{ddf,volo,involution,carafe} generate spatial-specific filters at each pixel via a bypath network. Taking Involution~\cite{involution} as a representative, it uses a $1 \times 1$ convolution to generate filters at each pixel and applies them to the neighboring area of that pixel. It can be written as
\begin{equation}
    \textbf{f}_i = \sum_{j \in \Theta}\textrm{Norm}(\textbf{x}_i \textbf{w}) \textbf{x}_j
\label{eq:dyfilter}
\end{equation}
where $\textbf{w}$ is the $1 \times 1$ convolutional weight of the filter generation branch. $\textrm{Norm}$ represents the normalization method applied to the generated filters, which is the identity mapping in Involution. For other cases, such as DDF~\cite{ddf} and Outlooker~\cite{volo}, the normalization methods are the filter normalization~\cite{ddf} and softmax, respectively.

\vspace{1mm}
\noindent\textbf{LSA.} Unlike dynamic filters, LSA uses attention maps of local windows as spatial-specific filters. In terms of formulation, LSA can be written as
\begin{equation}
    \textbf{f}_i = \sum_{j \in \Omega}\textrm{Softmax}_j(\textbf{q}_i \textbf{k}_j) \textbf{v}_j
\label{eq:lsa}
\end{equation}
where  $\textbf{q}_i, \textbf{k}_j, \textbf{v}_j$  are the query / key / value vectors at pixel $i$ and $j$, respectively. They are generated from the input feature via linear mappings. $\Omega$ notes the local window area. Here, we omit the multi-head strategy for simplicity.
 
\vspace{1mm}
\noindent\textbf{Unified paradigm.} To unify the above strategies, we first consider $\textbf{w}$ in Equation~\ref{eq:dyfilter} as a kind of relative position embedding, and consider $\textbf{w}_{j-i}$ in Equation~\ref{eq:dwconv} as a kind of relative position bias. Then, these spatial processing strategies can be unified into one paradigm, which can be written as
\begin{equation}
    \textbf{f}_i = \sum_{j \in \Phi}\textrm{Norm}_j(\textbf{q}_i \textbf{k}_j + \textbf{q}_i \textbf{r}^k_{j-i} + \textbf{r}^q_{j-i} \textbf{k}_j + \textbf{r}^b_{j-i}) \textbf{v}_j
    \label{eq:unify}
\end{equation}
where $\Phi$ can be either the local window $\Omega$ or the neighboring area $\Theta$; $\textrm{Norm}_j$ can be either the identity mapping, the filter normalization, or softmax; $\textbf{r}^k_{j-i}$ and $\textbf{r}^q_{j-i}$ are relative position embeddings, and $\textbf{r}^b_{j-i}$ denotes the relative position bias.

DwConv, dynamic filters, and LSA are all special cases of this unified paradigm. For example, when only using $\textbf{r}^b_{j-i}$ as parameterization, leveraging the identity mapping as $\textrm{Norm}_j$, and adopting the neighboring area $\Theta$ as $\Phi$, Equation~\ref{eq:unify} will degenerate to DwConv. Similarly, if the parameterization is set to $\textbf{q}_i \textbf{r}^k_{j-i}$, $\textrm{Norm}_j$ is set to the identity mapping, and $\Phi$ is set to the neighboring area $\Theta$, then Equation~\ref{eq:unify} becomes a variant of Involution, where $\textbf{r}^k_{j-i}$ is equivalent to $\textbf{w}$ in Equation~\ref{eq:dyfilter}. We can also get LSA from this paradigm by changing the parameterization, $\textrm{Norm}_j$, and $\Phi$. 
Therefore, the spatial processing of LSA and its counterparts are essentially different in three factors: \emph{parameterization}, \emph{normalization}, and \emph{filter application}. We then investigate each factor through extensive comparisons.

\subsection{Investigation}

\vspace{1mm}
\noindent\textbf{Parameterization.} We first compare the effect of different parameterizations by setting $\Phi$ as the local window and setting $\textrm{Norm}_j$ as softmax. The results are exhibited in Table~\ref{tbl:param}.
As can be seen, when applying filters to local windows, the parameterization strategy of dynamic filters (Net2) is better than that of the standard LSA (Net1). Also, a variant of dynamic filters (Net6) is on par with the LSA in ~\text{Swin-T}. Moreover, Net7 indicates that combining the parameterization of LSA with dynamic filters can further boost the performance, when applying filters to local windows.

\vspace{1mm}
\noindent\textbf{Normalization.} Normalization is another factor that influences spatial processing. DwConv and Involution~\cite{involution} adopt identity mapping as normalization. DDF~\cite{ddf} introduces the filter normalization. LSA and Outlooker~\cite{volo} employ the softmax function as normalization. We fairly compare these options under the same conditions. We choose the Net7 in Table~\ref{tbl:param} as the baseline, and only change the normalization part each time. As can be seen in Table~\ref{tbl:norm}, the identity mapping causes the training crash, and softmax is better than the filter normalization. This indicates that the normalization part should not be blamed for the mediocre performance of LSA. For more comparisons, please refer to the appendix.

\begin{table}[t]
    \small
    \addtolength{\tabcolsep}{-.5pt}
    \centering
    \begin{tabular}{lccccc}
        \toprule
        Model & $\textbf{q}_i \textbf{k}_j$ & $\textbf{q}_i \textbf{r}^k_{j-i}$ & $\textbf{r}^q_{j-i} \textbf{k}_j$ & $\textbf{r}^b_{j-i}$ & \text{Acc@1}\\
        \midrule
        \text{Swin-T$^{\dagger}$~\cite{swin}} & \checkmark & & & \checkmark & 81.3\\
        \midrule
        \text{Net1$^{\dagger}$~\cite{swin}} & \checkmark & & & & 80.1$_{\blue{--1.2}}$\\
        \text{Net2} &  & \checkmark & & & 80.9$_{\blue{--0.4}}$\\
        \text{Net3} & & & \checkmark &  & 80.9$_{\blue{--0.4}}$\\
        \text{Net4} & & & & \checkmark  & 79.8$_{\blue{--1.5}}$\\
        \text{Net5} & & \checkmark & \checkmark &  & 81.1$_{\blue{--0.2}}$\\
        \text{Net6} & & \checkmark & \checkmark & \checkmark & 81.3\hspace{1.1mm}$_{\green{0.0}}$ \\
        \text{Net7} & \checkmark & \checkmark & \checkmark & \checkmark & 81.8$_{\red{+0.5}}$\\
		\bottomrule
    \end{tabular}
    \caption{\textbf{Comparison of different parameterization strategies.} Swin-T~\cite{swin} is chosen as the baseline. Acc@1 means the top-1 accuracy. $\dagger$~denotes the results are reported in~\cite{swin}.}
    \label{tbl:param}
\end{table}

\begin{table}[t]
    \small
    \addtolength{\tabcolsep}{-.5pt}
    \centering
    \begin{tabular}{lcccc}
        \toprule
        Model & Identity & \text{Filter Norm} & Softmax & \text{Acc@1}\\
        \midrule
        \multirow{3}{*}{Net7} & \checkmark & & & crash\\
                              & & \checkmark & & 81.4\\
                              & & & \checkmark & 81.8\\
		\bottomrule
    \end{tabular}
    \caption{\textbf{Comparison of different normalization strategies.} We only switch the normalization of the Net7 in for fair comparison.}
    \label{tbl:norm}
\end{table}

\begin{table}[t]
    \small
    \addtolength{\tabcolsep}{-.5pt}
    \centering
    \begin{tabular}{lccc}
        \toprule
        Model & \text{Local Window} & {Neighboring} & \text{Acc@1}\\
        \midrule
        \multirow{2}{*}{Swin-T$^{\dagger}$~\cite{swin}} & \checkmark & & 81.3\\
                                            & & \checkmark & 81.4\\
        \midrule
        \multirow{2}{*}{Net6} & \checkmark & & 81.3\\
                              & & \checkmark & 82.0\\
        \midrule
        \multirow{2}{*}{Net7} & \checkmark & & 81.8\\
                              & & \checkmark  & \hspace{1.5mm}82.0$^*$\\
                              & & \checkmark  & \hspace{1.5mm}82.4$^\ddagger$\\
		\bottomrule
    \end{tabular}
    \caption{\textbf{Comparison of different filter applications.} We only switch the setting of the $\Phi$ in Equation~\ref{eq:unify} for fair comparison. $\dagger$~denotes the results are reported in~\cite{swin}. *~denotes that the model is implemented using SDC~\cite{vil} and trained on one node. $\ddagger$~denotes that the model is implemented using the unfold operation and trained on two nodes.}
    \label{tbl:application}
\end{table}

\vspace{1mm}
\noindent\textbf{Filter application.} The last factor of spatial processing is how filters are applied. LSA in Swin Transformer applies attention maps to non-overlapping local windows. In contrast, DwConv and dynamic filters apply filters to sliding neighboring areas. This difference can be described as using a different setting of $\Phi$ in Equation~\ref{eq:unify}. We choose Swin-T, Net6, and Net7 as the baseline. Following the control variable principle, we only switch $\Phi$ in the first three stages of those models.
We implement the neighboring case Net7 (noted as Net7-N) in two ways. We use SDC~\cite{vil} to implement it for saving GPU memory, so that we can use one node (8 GPUs with 32G GPU memory) to train Net7-N like others. We also implemented the unfold version, but which consumes huge GPU memories and requires multiple nodes to train. Table~\ref{tbl:application} shows the comparison results. When applying filters to neighboring areas, both Net6 and Net7 get significantly improved, indicating that the neighboring application is critical for spatial processing.
\subsection{Discussion}

\vspace{1mm}
\noindent\textbf{Key factors.} Based on the above investigations, the factors that make LSA mediocre can be summarized in two folds: one key factor affecting performance is relative position embeddings. Even when applying filters to local windows, variants (Net5 and Net6) with relative position embedding $\textbf{r}^k_{j-i}$ and $\textbf{r}^q_{j-i}$ achieve similar performance as Swin-T. Net7 further surpasses Swin-T by 0.5\% on top-1 accuracy. The other critical factor is the filter application. Applying filters on query-centered neighboring areas considerably boosts the performance of Net6 and Net7. So far, we can empirically answer the question we raised at beginning. The reason why DwConv can match the performance of LSA is because of the neighboring filter application. Without that, DwConv degenerates to a variant of Net4, which is inferior to LSA. Similarly, the reason why dynamic filters perform better than LSA is because of the relative position embedding and the neighboring filter application. Integrating these factors into LSA (Net7-N) achieves the best performance among all these variants.

\vspace{1mm}
\noindent\textbf{Local window v.s. Neighboring.} The peak performance of the local window version is worse than the neighboring one. Another disadvantage of the local window is that it requires strategies like window shifting to exchange information between windows, which limits the model design to have pairs of layers at each stage.

There is no such thing as a free lunch. The drawback of the neighboring version is low throughput. It is not easy to calculate the dot product between quires and keys in sliding neighboring areas. It requires sliding chunk~\cite{sdc}, unfold operations, or specialized CUDA implementations, which are either memory-consuming or time-consuming. How to avoid the dot product while maintaining good performance is a challenging problem.

\section{Enhanced Local Self-Attention}

\begin{figure}[t]
\centering 
\includegraphics[width=.95\linewidth]{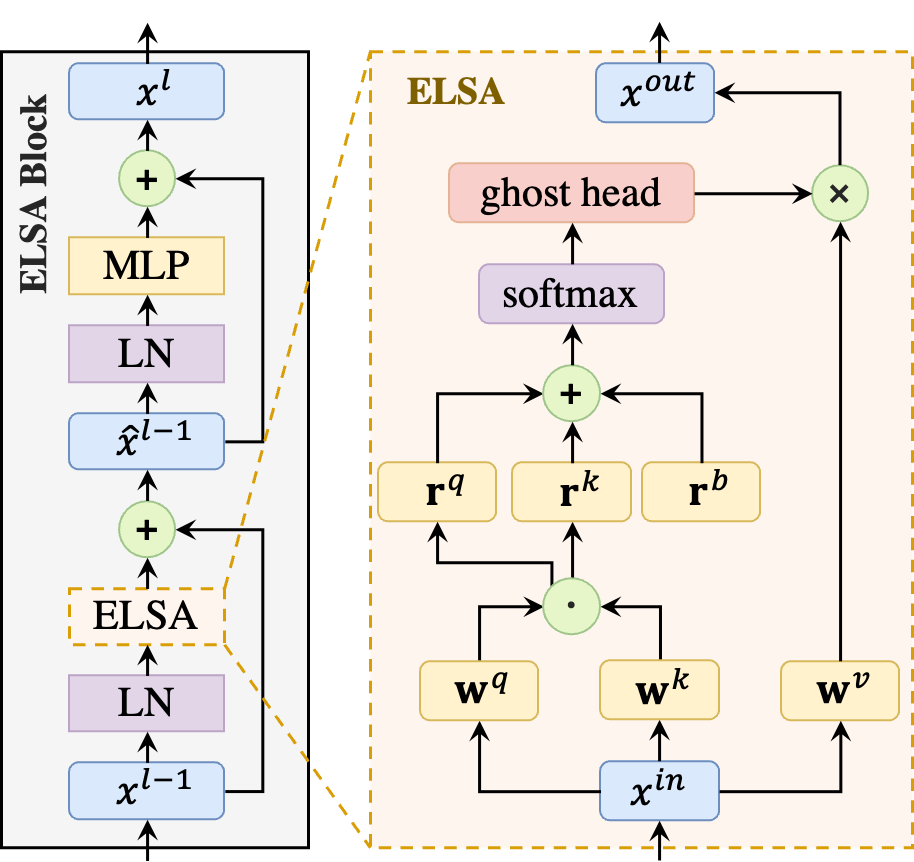}
\caption{\textbf{Illustration of the ELSA block.} ELSA can seamlessly replace LSA or dynamic filters in models.} 
\label{fig:elsa} 
\end{figure}

In addition to answering the question we raised, more importantly, we design a new local self-attention module that surpasses both the LSA in Swin Transformer and dynamic filters. This is accomplished with our enhanced local self-attention (ELSA), where the key techniques are Hadamard attention and the ghost head module. 
In terms of formulation, ELSA can be written as
\begin{equation}
    \textbf{f}_i = \sum_{j \in \Theta}\textrm{G}(\textbf{h}_{j-i}) \textbf{v}_j
    \label{eq:net6}
\end{equation}
where $\textbf{h}_{j-i}$ is the Hadamard attention value, $\textrm{G}(\cdot)$ notes the ghost head mapping, $\textbf{f}_i$ is the output feature at pixel $i$, and $\textbf{v}_j$ is the value vector at pixel $j$. Figure~\ref{fig:elsa} illustrates the overall structure of ELSA.

\vspace{1mm}
\noindent\textbf{Hadamard attention.}
First, we review Net7-N and Net6-N, which get the highest and the second-highest accuracy in our investigation. As discussed above, Net7-N brings in difficulties on implementation and inference. In the neighboring case, $\textbf{q}_i\textbf{k}_j$ in Net7-N needs to be implemented using unfold, sliding chuncks~\cite{vil}, or CUDA operations, which are either memory-consuming or time-consuming. As shown in Equation~\ref{eq:net6}, Net6-N removes the dot product term, thus getting rid of difficulties in filter / attention generation.

\begin{equation}
    \textbf{f}_i = \sum_{j \in \Theta}\textrm{Softmax}_j(\textbf{q}_i \textbf{r}^k_{j-i} + \textbf{r}^q_{j-i} \textbf{k}_j + \textbf{r}^b_{j-i}) \textbf{v}_j
    \label{eq:net6}
\end{equation}

However, the performance of Net6-N is slightly worse than Net7-N.  We find that the lower accuracy of Net6-N may be due to the lower mapping order. Specifically, Net-7 can be considered as a third-order mapping of the input $\textbf{x}$, because Net-7 contains the second-order term $\textbf{q}_i \textbf{k}_j$ and combine it with the value $\textbf{v}$. However, since Net6-N removes the dot product $\textbf{q}_i \textbf{k}_j$, which becomes a second-order mapping of the input $\textbf{x}$. 

One common hypothesis in deep learning is that the mappings with the higher order have stronger fitting ability~\cite{linming1}.
Thus, we want to design a module that maintains the third-order mapping just like Net7-N, but without using the dot product of the query and key matrices. To accomplish this, we propose Hadamard attention which introduces Hadamard product between queries and keys as the second-order term. In terms of formulation, Hadamard attention can be written as
\begin{equation}
\small
    \textbf{h}_{j-i} = \textrm{Softmax}_j((\textbf{q}_i\odot\textbf{k}_i) \textbf{r}^k_{j-i} + \textbf{r}^q_{j-i} (\textbf{q}_j\odot\textbf{k}_j) + \textbf{r}^b_{j-i}))
    \label{eq:hadamard}
\end{equation}
where $\odot$ means Hadamard product. With this simple yet effective modification, Equation~\ref{eq:hadamard} becomes a third-order representation of the input $\textbf{x}$.
 
Besides being a third-order representation, Hadamard attention is easy to implement. Unlike Net7-N that needs memory- / time-consuming operations to calculate $\textbf{q}_i\textbf{k}_j$,  $\textbf{q}_i\odot\textbf{k}_i$ and $\textbf{q}_j\odot\textbf{k}_j$ can be implemented via simple element-wise multiplication of the query and key feature maps. Also, $\textbf{r}^k_{j-i}$ and $\textbf{r}^q_{j-i}$ are equivalent to 1 $\times$ 1 convolutional filters. Thus, Equation~\ref{eq:hadamard} can be implemented by feature multiplication followed by convolution layers. For more implementation details please refer to the appendix and our code.

\vspace{1mm}
\noindent\textbf{Ghost Head.} In Section~\ref{sec:channel_setting}, we reveal the relation between the number of heads and model performance. Although channel setting is not the most critical factor that affects performance, more heads with better channel capacity still slightly boost the performance of DwConv-Swin-T in Figure~\ref{fig:channel_settings}. As directly enlarging the head number of LSA brings in no gain, we provide another cheap yet effective way to enrich channel capacity. Motivated by GhostNet~\cite{ghostnet}, we propose the ghost head module which expands the original heads by combining them with two static matrices, represented as

\vspace{-3mm}
\begin{equation}
    \hat{h}_{j-i}^{c} 
    = \textrm{Pow}(o_{j-i}^{c}, \lambda) h_{j-i}^{c'} + \gamma s_{j-i}^{c}
    \label{eq:ghost_head}
\end{equation}
where $h_{j-i}^{c}$ / $\hat{h}_{j-i}^{c} \in \mathbb{R}$ are the Hadamard attention values before / after the ghost head module, which are at $c'$ / $c$-th channel and corresponds to the relative offset $j-i$; the relation between $c$ and $c'$ is $c' = c~\%~n_h$, where $n_h$ is the number of heads and $\%$ is the mod operation;
$O$, $S \in \mathbb{R}^{n_c \times ks\times ks}$ are two static matrices; $o_{j-i}^{c}$, $s_{j-i}^{c}$ are the values in $O$, $S$ at the $c$-th channel corresponding to the relative offset $j-i$; $\textrm{Pow}(\cdot)$ is the power operation; $\lambda$, $\gamma$ are two hyper-parameters.
The demo code of the ghost head is summarized in Algorithm~\ref{alg:ghost_head}. In the real implementation, we write CUDA operations to avoid large GPU memory consumption.

\begin{algorithm}[t]
\definecolor{codeblue}{rgb}{0.25,0.5,0.25}
\lstset{
	backgroundcolor=\color{white},
	basicstyle=\fontsize{7.2pt}{7.2pt}\ttfamily\selectfont,
	columns=fullflexible,
	breaklines=true,
	captionpos=b,
	commentstyle=\fontsize{7.2pt}{7.2pt}\color{codeblue},
	keywordstyle=\fontsize{7.2pt}{7.2pt},
}
\begin{lstlisting}[language=python]
# B: batch size, C: channel size
# N: the number of pixels
# H: the number of heads, K: kernel size
# h_attn: Hadamard attention with size (B, H, N, K*K)
# lambda, gamma: hyperparameters

def init()
    mul_matrix = nn.Parameters(torch.randn(C, K, K))
    add_matrix = nn.Parameters(torch.zeros(C, K, K))
    trunc_normal_(add_matrix, std=0.02)
    
def ghost_head(h_attn):
    # change the size of h_attn to (B, 1, H, N, K*K)
    h_attn = h_attn.unsqueeze(1)
    
    # reshape the size of matrices
    mul_matrix = mul_matrix.reshape(1, C//H, H, 1, K*K)
    add_matrix = add_matrix.reshape(1, C//H, H, 1, K*K)
    
    # Equation 8
    h_attn = (mul_matrix ** lambda) * h_attn + gamma * add_matrix
    return h_attn.reshape(B, C, N, K*K)
\end{lstlisting}
\caption{{Demo code of ghost head (PyTorch-like)}}
\label{alg:ghost_head}
\end{algorithm}

The ghost head is a cheap module. Its overhead is only $O(n_c\times ks\times ks \times n_p)$, where $n_c$ is the number of channels, $ks$ is the filter size (\ie the size of neighboring areas), and $n_p$ is the number of pixels. Recently, Refiner~\cite{refiner} is also proposed to adjust attention after softmax. Unlike them, the ghost head does not leverage heavy convolutions and linear mappings, but only uses two simple static matrices. Also, the main purpose of the ghost head is not to refine attention, but to enrich channel capacity.

\section{Experiments}
We evaluate our ELSA in the Swin Transformer and VOLO architectures on ImageNet-1K image classification~\cite{imagenet}, COCO object detection~\cite{coco}, and ADE20K semantic segmentation~\cite{ade20k}. For Swin Transformer~\cite{swin}, we drop-in replace LSA with ELSA without changing any architecture / hyperparameters. For VOLO~\cite{volo}, we adjust several hyperparameters of ELSA Blocks, while maintaining all other parts unchanged.

\subsection{Image Classification on ImageNet-1K}
\label{sec:expsetting}
\vspace{1mm}
\noindent\textbf{Settings.} Our major evaluations are conducted on the ImageNet-1K~\cite{imagenet} dataset. During training, no extra training images are used. \text{Our code is based on Pytorch~\cite{pytorch}}, timm~\cite{timm}, DDF~\cite{ddf}, Swin Transformer~\cite{swin}, and VOLO~\cite{volo}.  The detailed setups are as follows.

\vspace{1mm}
\noindent\emph{Swin Transformer.} We replace LSA blocks of the first three stages with our ELSA blocks, and train ELSA-Swin following~\cite{swin}. In particular, AdamW~\cite{adamw} is selected as the optimizer. The base learning rate is set to 1e-3, which is scaled following the linear strategy, \ie $lr = lr_{base} \times \frac{batch\_size}{1024}$, and decays following the cosine strategy. We train models for 310 epochs, where the first 20 epochs are used for warm-up, and the last 10 epochs are used for cool-down. The weight decay of 5e-2 is used. We also leverage the same augmentation and regularization strategies as~\cite{swin}. Exponential moving average (EMA)~\cite{ema, mean_teacher} is used in ELSA-Swin-B training. All models are trained / evaluated on $224 \times 224$ resolution unless otherwise specified.

\vspace{1mm}
\noindent\emph{VOLO.} We replace all Outlooker modules in VOLO with ELSA. We train ELSA-VOLOs following the training protocol of VOLO. Most settings of VOLO are similar to those of Swin Transformer, except for the following differences. The base learning rate is set to 1.6e-3 for VOLO-D1 and 8e-4 for VOLO-D5. The Token Labeling~\cite{lvvit} is used during training, thus, MixUp~\cite{mixup} and CutMix~\cite{cutmix} are replaced by MixToken~\cite{lvvit}. EMA is used in ELSA-VOLO-D5 training. Please refer to~\cite{volo} for more details.

\vspace{1mm}
\noindent\textbf{Ablation study.} We respectively analyze the effect of Hadamard attention and the ghost head module in ELSA. We choose Swin-T~\cite{swin} as our base architecture and experiment with different modifications to ELSA. 
Table~\ref{tbl:ablation} shows the results of ablation experiments. The performance of Swin-T is improved by 1.1\%, with only Hadamard attention. We observed another 0.3\% improvements by plugging-in the ghost head.

\begin{table}[t]
    \small
    \addtolength{\tabcolsep}{-1pt}
    \centering
    \begin{tabular}{ccccc}
        \toprule
        \text{HA} & \text{GH} & Params & FLOPs & \hspace{-2mm}\text{Acc@1}\\
        \midrule
        \multicolumn{2}{c}{\textit{Base Model}} & 28.3M & 4.5G & \hspace{-2mm}81.3\\
        \checkmark & & 29.0M & 4.7G & \hspace{2mm}82.4$_{\red{+1.1}}$\\
        \checkmark & \checkmark & \textbf{29.1M} & \textbf{4.8G} & \hspace{2mm}\textbf{82.7$_{\red{+1.4}}$}\\
		\bottomrule
    \end{tabular}
    \caption{\textbf{Evaluate different components of ELSA.} The Swin-T is chosen as the baseline. HA means Hadamard attention, and GH denotes the ghost head module.}
    \label{tbl:ablation}
\end{table}

Table~\ref{tbl:layer_types} further compares the performance between different types of layers on both Swin Transformer and VOLO. As can be seen, Swin-T and VOLO-D1 with ELSA respectively achieve 82.7\% and 84.7\% top-1 accuracy, which surpass other compared counterparts.

\begin{table}[t]
    \small
    \addtolength{\tabcolsep}{-1.8pt}
    \centering
    \begin{tabular}{llccc}
        \toprule
        Architecture & Layer type & Params & FLOPs &\text{Acc@1}\\
        \midrule
        \multirow{5}{*}{Swin-T~\cite{swin}} & LSA~\cite{swin} & 28M & 4.5G & 81.3\\
                                & DwConv~\cite{demystifying} & 24M & 3.7G & 81.6\\
                                & \text{D-DwConv}~\cite{demystifying} & 51M & 3.8G & 81.9\\
                                & DDF~\cite{ddf} & 46M & 3.9G & 82.0\\
                                & \textbf{ELSA} & \textbf{29M} & \textbf{4.8G} & \textbf{82.7}\\
        \midrule
        \multirow{4}{*}{VOLO-D1~\cite{volo}} & LSA~\cite{volo} & 27M & -- & 83.8\\
                                 & DwConv~\cite{volo} & 27M & -- & 83.8\\
                                 & Outlooker~\cite{volo} & 27M & 7.1G & 84.2\\
                                 & \textbf{ELSA} & \textbf{27M} & \textbf{8.0G} & \textbf{84.7}\\
		\bottomrule
    \end{tabular}
    \caption{\textbf{Comparison of different layers.} Consistently boosts baselines and surpasses compared counterparts, while using overhead similar to LSA.}
    \label{tbl:layer_types}
    \vspace{-3mm}
\end{table}

\vspace{1mm}
\noindent\textbf{Compare with the state-of-the-art models.}
We compare ELSA-Swin and ELSA-VOLO with other state-of-the-art models in Table~\ref{tbl:sota}.
\#Res represents resolutions used in validating / finetuning.  For fair comparisons, results are split into groups according to the number of parameters.

\begin{table}[t]
    \small
    \addtolength{\tabcolsep}{-2pt}
    \centering
    \begin{tabular}{lcccc}
        \toprule
        Model & Params & FLOPs & \#Res & \text{Acc@1}\\
        \midrule 
        T2T-ViT-14~\cite{t2t}       & 22M & 5.2G & 224$^2$ & 81.5\\
        CoAtNet-0~\cite{coatnet}    & 25M & 4.2G & 224$^2$ & 81.6\\
        Twins-SVT-S~\cite{twins}    & 24M & 2.9G & 224$^2$ & 81.7\\
        Swin-T~\cite{swin}          & 28M & 4.5G & 224$^2$ & 81.3\\
        VOLO-D1~\cite{volo}         & 27M & 7.1G & 224$^2$ & 84.2\\
        VOLO-D1$_{\uparrow384}$~\cite{volo}  & 27M & 20.8G & 384$^2$ & 85.2\\
        \textbf{ELSA-Swin-T}   & \textbf{28M} & \textbf{4.8G} & \textbf{224$^2$} & \textbf{82.7}\\
        \textbf{ELSA-VOLO-D1}  & \textbf{27M} & \textbf{8.0G} & \textbf{224$^2$} & \textbf{84.7}\\
        \textbf{ELSA-VOLO-D1$_{\uparrow384}$} & \textbf{27M} & \textbf{23.3G} & \textbf{384$^2$} & \textbf{85.7}\\
        \midrule
        T2T-ViT-24~\cite{t2t}       & 64M & 15.0G & 224$^2$ & 82.6\\
        CoAtNet-1~\cite{coatnet}    & 42M & 8.4G & 224$^2$ & 83.3\\
        Twins-SVT-B~\cite{twins}    & 56M & 8.6G & 224$^2$ & 83.2\\
        Swin-S~\cite{swin}          & 50M & 8.7G & 224$^2$ & 83.0\\
        \textbf{\text{ELSA-Swin-S}} & \textbf{53M} & \textbf{9.6G} & \textbf{224$^2$} & \textbf{83.5}\\
        \midrule
        CoAtNet-2~\cite{coatnet}    & 75M & 15.7G & 224$^2$ & 84.1\\
        Twins-SVT-L~\cite{twins}    & 99M & 15.1G & 224$^2$ & 83.7\\
        Swin-B~\cite{swin}          & 88M & 15.4G & 224$^2$ & 83.5\\
        VOLO-D3~\cite{volo}         & 86M & 20.9G & 224$^2$ & 85.4\\
        VOLO-D3$_{\uparrow448}$~\cite{volo} & 86M & 92.9G & 448$^2$ & 86.3\\
        \textbf{\text{ELSA-Swin-B}} & \textbf{93M} & \textbf{16.7G} & \textbf{224$^2$} & \textbf{84.0}\\
        \textbf{\text{ELSA-VOLO-D3}} & \textbf{87M} & \textbf{22.3G} & \textbf{224$^2$} & \textbf{85.7}\\
        \textbf{\text{ELSA-VOLO-D3$_{\uparrow448}$}} & \textbf{87M} & \textbf{98.6G} & \textbf{448$^2$} & \textbf{86.6}\\
        \midrule
        CoAtNet-3~\cite{coatnet}    & 168M & 34.7G & 224$^2$ & 84.5\\
        VOLO-D4~\cite{volo}         & 193M & 44.6G & 224$^2$ & 85.7\\
        VOLO-D4$_{\uparrow448}$~\cite{volo} & 193M & 194G & 448$^2$ & 86.8\\
        \midrule
        CaiT-M36$_{\uparrow448}$~\cite{cait} & 271M & 248G & 448$^2$ & 86.3\\
        CaiT-M48$_{\uparrow448}$~\cite{cait} & 356M & 330G & 448$^2$ & 86.5\\
        VOLO-D5~\cite{volo}         & 295M & 72.7G & 224$^2$ & 86.1\\
        VOLO-D5$_{\uparrow512}$~\cite{volo} & 295M & 407G & 512$^2$ & 87.1\\
        \textbf{\text{ELSA-VOLO-D5}} & \textbf{298M} & \textbf{78.5G} & \textbf{224$^2$} & \textbf{86.3}\\
        \textbf{\text{ELSA-VOLO-D5$_{\uparrow512}$}} & \textbf{298M} & \textbf{437G} & \textbf{512$^2$} & \textbf{87.2}\\
		\bottomrule
    \end{tabular}
    \caption{\textbf{Comparison of different backbones on the ImageNet-1K.} Simply plugging-in ELSA achieves state-of-the-art performance.}
    \vspace{-3mm}
    \label{tbl:sota}
\end{table}

As can be seen, for different model sizes, our proposed ELSA consistently boosts Swin Transformer and VOLO, while introducing little overhead. In particular, ELSA improves Swin-T, Swin-S, and Swin-B by 1.4\%, 0.5\%, and 0.5\%, where ELSA-Swin-S is comparable to the original Swin-B with two-thirds of parameters and FLOPs. Testing on the resolution of 224, ELSA-VOLO-D1 and ELSA-VOLO-D3 yield 84.7\% and 85.7\% top-1 accuracy, respectively. The performance of ELSA-VOLO-D3 matches the performance of the original VOLO-D4. However, VOLO-D4 costs more than $2\times$ parameters against ELSA-VOLO-D3. Without additional images, it is very difficult to improve the accuracy of large models under supervised training due to over-fitting. Even so, ELSA still slightly improves VOLO-D5.

It is worth noting that all these records of ELSA-Swin are obtained without modifying any hyperparameters in Swin Transformers, which may not be the optimal setting for our ELSA. For VOLO, we also keep the structure and hyperparameters of other parts unchanged. Under such control variable principle, the use of ELSA blocks still achieves state-of-the-arts. Redesigning the macro architecture / hyperparameters manually or by NAS may yield better Pareto performance.

\subsection{Object Detection on COCO}

\vspace{1mm}
\noindent\textbf{Settings.} Object detection and instance segmentation experiments are conducted on the COCO dataset. We report the performance on the validation subset, and use the mean average precision (AP) as the metric. 
We evaluate ELSA-Swin in Mask RCNN / Cascade Mask RCNN~\cite{cascade, mask_rcnn}, which is a common practice in~\cite{vil, focal_attn, pvt, wang2021crossformer, region_vit}. Following the common training protocol, we apply multi-scale training, scaling the shorter side of the input from 480 to 800 while keeping the longer side no more than 1333. AdamW~\cite{adamw} is adopted as the optimizer with an initial learning rate of 1e-4, weight decay of 5e-2, and batch size of 16. 
For fair comparisons, all backbones are pretrained using the ImageNet-1K only, and finetuned on the COCO with 1$\times$ schedule (12 epochs). 
Our implementation is based on Swin Transformer~\cite{swin} and mmdetection~\cite{mmdet}. 

\begin{table}[t]
    \small
    \addtolength{\tabcolsep}{-4.8pt}
    \centering
    \begin{tabular}{c|ccc|ccc|cc}
        \toprule
        \multicolumn{9}{c}{Mask RCNN with 1$\times$ schedule}\\
        \hline
        Backbone & AP$^{b}$ & AP$_{50}^{b}$ & AP$_{75}^{b}$ & AP$^{m}$ & AP$_{50}^{m}$ & AP$_{75}^{m}$ & Params & FLOPs\\
        \hline  
        PVT-M~\cite{pvt} & 42.0 & 64.4 & 45.6 & 39.0 & 61.6 & 42.1 & 64M & --\\
        Region-S~\cite{region_vit} & 44.2 & 67.3 & 48.2 & 40.8 & 64.1 & 44.0 & 51M & 183G\\
        Focal-T~\cite{focal_attn} & 44.8 & -- & -- & 41.0 & -- & -- & 49M & 291G\\
        ViL-S~\cite{vil} & 44.9 & 67.1 & 49.3 & 41.0 & 64.2 & 44.1 & 45M & 218G\\
        Swin-T~\cite{swin} & 43.7  & 66.6 & 47.7 & 39.8 & 63.3 & 42.7 & 48M & 267G\\
        \textbf{ELSA-T} & \textbf{45.6}  & \textbf{67.9} & \textbf{50.3} & \textbf{41.1} & \textbf{64.8} & \textbf{44.0} & \textbf{49M} & \textbf{269G}\\
        \hline  
        PVT-L~\cite{pvt} & 42.9 & 65.0 & 46.6 & 39.5 & 61.9 & 42.5 & 81M & --\\
        Region-B~\cite{region_vit} & 46.3 & 69.1 & 51.2 & 42.4 & 66.2 & 45.6 & 93M & 464G \\
        Focal-S~\cite{focal_attn} & 47.4 & -- & -- & 42.8 & -- & -- & 71M & 401G\\
        ViL-M~\cite{vil} & 47.6 & 69.8 & 52.1 & 43.0 & 66.9 & 46.6 & 60M & 294G\\
        Swin-S$^\dagger$~\cite{swin}  & 46.5 & -- & -- & 42.1 & -- & -- & 69M & 354G\\
        \textbf{ELSA-S}  & \textbf{48.3}  & \textbf{70.4} & \textbf{52.9} & \textbf{43.0} & \textbf{67.3} & \textbf{46.4} & \textbf{72M} & \textbf{367G}\\
        \midrule
        \multicolumn{9}{c}{Cascade Mask RCNN with 1$\times$ schedule}\\
        \hline  
        Backbone & AP$^{b}$ & AP$_{50}^{b}$ & AP$_{75}^{b}$ & AP$^{m}$ & AP$_{50}^{m}$ & AP$_{75}^{m}$ & Params & FLOPs\\
        \hline  
        Swin-T~\cite{swin} & 48.1  & 67.1  & 52.2 & 41.7 & 64.4 & 45.0 & 86M & 745G\\
        \textbf{ELSA-T} & \textbf{49.8}  & \textbf{68.9} & \textbf{54.2} & \textbf{43.0} & \textbf{66.1} & \textbf{46.3} & \textbf{86M} & \textbf{748G}\\
        \hline  
        Swin-S~\cite{swin}  & 50.3  & 69.7 & 54.4 & 43.4 & 67.0 & 46.7 & 107M & 838G\\
        \textbf{ELSA-S}  & \textbf{51.6}  & \textbf{70.5} & \textbf{56.0} & \textbf{44.4} & \textbf{67.8} & \textbf{47.8} & \textbf{110M} & \textbf{846G}\\
        \bottomrule
    \end{tabular}
    \caption{\textbf{Comparison of different backbones on the COCO validation set.} AP$^{b}$ / AP$^{m}$ denote the mean average precision of detection / segmentation. $\dagger$~denotes the results are reported in~\cite{focal_attn}.}
    \label{tbl:coco}
    \vspace{-2mm}
\end{table}

\vspace{1mm}
\noindent\textbf{Results.} Table~\ref{tbl:coco} lists experimental results. As can be seen, ELSA-Swin-T and ELSA-Swin-S (noted as ELSA-T / ELSA-S) respectively improve the corresponding baseline by 1.9 AP and 1.8 AP in detection, both outperforming other methods within their group. Note that, unlike ViL~\cite{vil} and RegionViT~\cite{region_vit}, ELSA-Swin does not modify the macro architecture / hyperparameters of Swin Transformer.
Cascade Mask RCNNs with ELSA-Swin-T and ELSA-Swin-S achieve 49.8 AP and 51.6 AP in detection, which are 1.7 AP and 1.3 AP higher than their baselines. The appendix shows more comparisons with 3$\times$ scheduleon the COCO dataset.

\subsection{Semantic Segmentation on ADE20K}

\vspace{1mm}
\noindent\textbf{Settings.} We evaluate the semantic segmentation performance of ELSA-Swin on the ADE20K~\cite{ade20k}, which contains 20K training, 2K validation, and 3K testing images, covering 150 semantic categories. Following~\cite{swin, demystifying, volo}, UperNet~\cite{upernet} is selected as the baseline framework. During training, AdamW is adopted as the optimizer. The initial learning rate is set to 6e-5, weight decay is set to 1e-2. All models are trained for 160K iterations with linear learning rate decay, and a linear warmup of 1500 iterations. We use default augmentation settings in mmsegmentation~\cite{mmseg} where the resolution of the input is set to 512 $\times$ 512. During inference, we perform multi-scale test with interpolation rates of [0.75, 1.0, 1.25, 1.5, 1.75]. For more details, please refer to~\cite{swin, mmseg} and our code.

\begin{table}[t]
    \small
    \addtolength{\tabcolsep}{-1pt}
    \centering
    \begin{tabular}{lcccc}
        \toprule
        Backbone  & Params & FLOPs & \hspace{-.33mm}MS mIoU\\
        \midrule
        Swin-T~\cite{swin} & 60M & 945G & \hspace{-.33mm}45.8 \\
        Focal-T~\cite{focal_attn} & 62M & 998G & \hspace{-.33mm}47.0 \\
        Twins-SVT-S~\cite{twins} & 54M & -- & \hspace{-.33mm}47.1\\
        \textbf{ELSA-Swin-T} & \textbf{61M} & \textbf{946G} & \hspace{3.67mm}\textbf{47.7$_{\red{+1.9}}$}\\
        \midrule
        Swin-S~\cite{deit} & 81M & 1038G & \hspace{-.33mm}49.5 \\
        Focal-S~\cite{focal_attn} & 85M & 1130G & \hspace{-.33mm}50.0 \\
        Twins-SVT-B~\cite{twins} & 89M & -- & \hspace{-.33mm}48.9 \\
        \textbf{ELSA-Swin-S} & \textbf{85M} & \textbf{1046G} & \hspace{3.67mm}\textbf{50.3$_{\red{+0.8}}$}\\
        \midrule
        Swin-B~\cite{swin} & 121M & 1188G & 49.7 \\
        Focal-B~\cite{focal_attn} & 126M & 1354G  & 50.5\\
        Twins-SVT-L~\cite{twins} & 133M & -- & 50.2\\
		\bottomrule
    \end{tabular}
    \caption{\textbf{Comparison of different backbones on the ADE20K validation set.} UperNet~\cite{upernet} is adopted as the framework. All compared backbones are pretrained with the ImageNet-1K only. }
    \label{tbl:ade20k}
    \vspace{-2mm}
\end{table}

\vspace{1mm}
\noindent\textbf{Results.} Table~\ref{tbl:ade20k} shows the mean IoU with multi-scale testing (MS mIoU), model size (Param), and FLOPs of different methods. Results are split into three groups based on the number of model parameters. For fair comparisons, all compared backbones are pretrained using the ImageNet-1K only. UperNet with ELSA-Swin-T is 1.9 higher on MS mIoU than the Swin-T version. Adopting ELSA-Swin-S as the backbone achieves 50.3 MS mIoU, which is 0.8 higher on MS mIoU than Swin-S, and is even better than Swin-B and Twins-SVT-L in the third group.

\section{Conclusions}

In this work, we investigate LSA and its counterparts in detail from channel settings and spatial processing to empirically understand why LSA performs mediocre.
It is revealed that the relative position embedding and the neighboring filter application are critical reasons why DwConv and dynamic filters perform similar or better than LSA. Based on these observations, we further propose the enhanced local self-attention (ELSA) with Hadamard Attention and the ghost head, which can seamlessly replace LSA and its counterparts in various networks. Experiments show that, without other architecture / hyperparameter modifications, ELSA can consistently improve the baseline, regardless of the model size and tasks, with little overhead being introduced.

\section*{Acknowledgements}
This work was supported by Alibaba Group through Alibaba Research Intern Program and the National Natural Science Foundation of China (No.61976094).

{\small
\bibliographystyle{ieee_fullname}
\bibliography{references}
}

\appendix
\section{Additional Investigation}
\label{sec:invest_supp}

In addition to peak performance, we also compared the convergence speed between Filter Normalization~\cite{ddf} and softmax. We select Net7 as our baseline and show the accuracy at different epochs in Table~\ref{tbl:convergence}.
As can be seen, the Filter Normalization can speed up model convergence at early epochs. However, due to the constraint of 0 mean and 1 variance in Filter Normalization, the flexibility of filters are limited. Therefore, in the last 100 epochs, Net7 with Filter Normalization lags behind the softmax version.


\section{COCO Detection with 3$\times$ Schedule}
Table~\ref{tbl:coco_3x} lists the COCO detection results with 3$\times$ schedule (36 epochs). As can be seen, ELSA-Swin-T (noted as ELSA-T) improves the baseline by 1.5 box AP / 1.1 mask AP; ELSA-Swin-S (noted as ELSA-S) boosts the baseline by 0.7 box AP / 0.3 mask AP. Both of their box AP surpass other counterparts within their group. Cascade Mask R-CNNs with ELSA-Swin-T and ELSA-Swin-S achieve 51.1 AP and 52.3 AP in detection, which are 0.6 AP and 0.5 AP higher than their baselines. It is worth noting that the ELSA-Swin-S version even surpasses the the Swin-B one.

\section{Ghost Head on Global Self-Attention}
Beyond ELSA, we also evaluate the performance of the ghost head when applied to global self-attention. We chose DeiT-Ti~\cite{deit} as the baseline. We first remove the class / distillation tokens in DeiT-Ti and use the global average pooling to generate the feature for classification. Then, we apply the ghost head module to global self-attention maps after softmax. Instead of expanding the number of heads to the number of channels, in the global self-attention case, the ghost head module only doubles the number of heads. As can be seen in Table~\ref{tbl:global_attn}, the ghost head module also considerably improves the the baseline by 3.7\% on top-1 accuracy.

\section{Implementation of Hadamard Attention}
We do not implement Hadamard attention strictly according to Equation~\ref{eq:hadamard}. Hadamard attention is implemented as a variant of Equation~\ref which is faster and better. Here, we will show how to get our implementation step by step from Equation~\ref{eq:hadamard} in the form of pseudo code.

\begin{table}[t]
    \addtolength{\tabcolsep}{-.5pt}
    \small
    \centering
    \begin{tabular}{lcccccc}
        \toprule
        Epochs & 50 & 100 & 150 & 200 & 250 & 300 \\
        \midrule
        Filter Norm~\cite{ddf} & 71.5 & 74.9 & 77.2 & 79.4 & 80.8 & 81.4 \\
        Softmax & 70.3 & 74.7 & 77.1 & 79.3 & 81.0 & 81.8 \\                    
		\bottomrule
    \end{tabular}
    \caption{\textbf{Top-1 Accuracy at different training epochs.} Filter normalization speeds up the model convergence at early epochs, but falls into the local minimum due to constraints.}
    \label{tbl:convergence}
\end{table}

\begin{table}[t]
    \small
    \addtolength{\tabcolsep}{-4.8pt}
    \centering
    \begin{tabular}{c|ccc|ccc|cc}
        \toprule
        \multicolumn{9}{c}{Mask RCNN with 3$\times$ schedule}\\
        \hline
        Backbone & AP$^{b}$ & AP$_{50}^{b}$ & AP$_{75}^{b}$ & AP$^{m}$ & AP$_{50}^{m}$ & AP$_{75}^{m}$ & Params & FLOPs\\
        \hline  
        PVT-M~\cite{pvt} & 44.2 & 66.0 & 48.2 & 40.5 & 63.1 & 43.5 & 64M & --\\
        Focal-T~\cite{focal_attn} & 47.2 & 69.4 & 51.9 & 42.7 & 66.5 & 45.9 & 49M & 291G\\
        ViL-S~\cite{vil} & 47.1 & 68.7 & 51.5 & 42.7 & 65.9 & 46.2 & 45M & 218G\\
        Swin-T~\cite{swin} & 46.0  & 68.1 & 50.3 & 41.6 & 65.1 & 44.9 & 48M & 267G\\
        \textbf{ELSA-T} & \textbf{47.5}  & \textbf{69.1} & \textbf{52.3} & \textbf{42.7} & \textbf{66.3} & \textbf{45.9} & \textbf{49M} & \textbf{269G}\\
        \hline  
        PVT-L~\cite{pvt} & 44.5 & 66.0 & 48.3 & 40.7 & 63.4 & 43.7 & 81M & --\\
        Focal-S~\cite{focal_attn} & 48.8 & 70.5 & 53.6 & 43.8 & 67.7 & 47.2 & 71M & 401G\\
        ViL-M~\cite{vil} & 48.9 & 70.3 & 54.0 & 44.2 & 67.9 & 47.7 & 60M & 294G\\
        Swin-S$^\dagger$~\cite{swin}  & 48.5 & 70.2 & 53.5 & 43.3 & 67.3 & 46.6 & 69M & 354G\\
        \textbf{ELSA-S}  & \textbf{49.2}  & \textbf{70.3} & \textbf{54.3} & \textbf{43.6} & \textbf{67.4} & \textbf{46.8} & \textbf{72M} & \textbf{367G}\\
        \midrule
        \multicolumn{9}{c}{Cascade Mask RCNN with 3$\times$ schedule}\\
        \hline  
        Backbone & AP$^{b}$ & AP$_{50}^{b}$ & AP$_{75}^{b}$ & AP$^{m}$ & AP$_{50}^{m}$ & AP$_{75}^{m}$ & Params & FLOPs\\
        \hline  
        Swin-T~\cite{swin} & 50.5 & 69.3 & 54.9 & 43.7 & 66.6 & 47.1 & 86M & 745G\\
        \textbf{ELSA-T} & \textbf{51.1}  & \textbf{69.7} & \textbf{55.3} & \textbf{44.2} & \textbf{67.2} & \textbf{48.1} & \textbf{86M} & \textbf{748G}\\
        \hline  
        Swin-S~\cite{swin}  & 51.8 & 70.4 & 56.3 & 44.7 & 67.9 & 48.5 & 107M & 838G\\
        \textbf{ELSA-S}  & \textbf{52.3}  & \textbf{70.9} & \textbf{57.1} & \textbf{45.2} & \textbf{68.4} & \textbf{49.2} & \textbf{110M} & \textbf{846G}\\
        \hline
        Swin-B~\cite{swin}  & 51.9  & 70.9 & 56.5 & 45.0 & 68.4 & 48.7 & 145M & 982G\\
        \bottomrule
    \end{tabular}
    \caption{\textbf{Comparison of different backbones on the COCO validation set.} AP$^{b}$ / AP$^{m}$ denote the mean average precision of detection / segmentation. $\dagger$~denotes the results are reported in~\cite{focal_attn}.}
    \label{tbl:coco_3x}
    \vspace{-2mm}
\end{table}

\begin{table}[t]
    \addtolength{\tabcolsep}{-2pt}
    \small
    \centering
    \begin{tabular}{lcccc}
        \toprule
        Models & Params & \#Res & Acc@1 & Acc@5\\
        \midrule
        DeiT-Ti Original~\cite{deit} & 6M & 224$^2$ & 72.2& 91.1\\
        DeiT-Ti GAP  & 6M & 224$^2$ & 73.8 & 92.2\\
        DeiT-Ti GAP w/ GH  & 6M & 224$^2$ & 77.5 & 93.7\\
		\bottomrule
    \end{tabular}
    \caption{\textbf{Evaluation of applying ghost head to global self-attention.} GAP means global average pooling, GH denotes the ghost head module.}
    \label{tbl:global_attn}
\end{table}

The strict implementation of Equation~\ref{eq:hadamard} is shown in Algorithm~\ref{alg:strict}, where the unfold operation is used to calculate $\textbf{q}_j\odot \textbf{k}_j$. The unfold operation is memory-consuming, while the main function of it is to shift the feature map and align the pixel $i$ and $j$, so that we can directly add $\textbf{q}_i\odot\textbf{k} _i$ with $\textbf{ q}_j\odot \textbf{k}_j$. Instead of using unfold operation, we can use depth-wise convolution to shift feature maps, which is shown in Algorithm~\ref{alg:equivalent1}.

Algorithm~\ref{alg:equivalent1} can be further simplified by merge two 1$\times$1 convolution layers to one layer, as shown in Algorithm~\ref{alg:equivalent2}. Note that Algorithm~\ref{alg:strict}, Algorithm~\ref{alg:equivalent1}, and Algorithm~\ref{alg:equivalent2} are completely equivalent. One can get the same output by setting the same model parameters. 

As can be seen in Algorithm~\ref{alg:equivalent2}, Equation~\ref{eq:hadamard} can be implemented by feature multiplication followed with a 1$\times$1 convolution and a group convolution. For our final implementation (a variant of Equation~\ref{eq:hadamard}), we slightly modify the sequence of these convolutional layers and add an activation function (GELU) between them, which is shown in the attached code.

\begin{algorithm}[t]
\definecolor{codeblue}{rgb}{0.25,0.5,0.25}
\lstset{
	backgroundcolor=\color{white},
	basicstyle=\fontsize{7.2pt}{7.2pt}\ttfamily\selectfont,
	columns=fullflexible,
	breaklines=true,
	captionpos=b,
	commentstyle=\fontsize{7.2pt}{7.2pt}\color{codeblue},
	keywordstyle=\fontsize{7.2pt}{7.2pt},
}
\begin{lstlisting}[language=python]
# B: batch size, C: channel size
# H, W: the height / width of feature map
# G: the number of heads, K: kernel size
# q, k: queries and keys in shape (B, C, H, W)

def init()
    rq = nn.Parameters(torch.randn(C, G, K*K))
    trunc_normal_(rq, std=0.02)
    rk = nn.Parameters(torch.zeros(C, G, K*K))
    trunc_normal_(rk, std=0.02)
    rb = nn.Parameters(G, K*K)
    trunc_normal_(rb, std=0.02)
    unfold_op = nn.Unfold(kernel_size=window_size, 
            padding=window_size//2, stride=1)
    
def cal_hp_rk(hp):
    return torch.einsum('bchw,cgk->bgkhw', hp, rk)

def cal_rq_hp(hp):
    hp = unfold_op(hp).reshape(B, C, K*K, H, W)
    return torch.einsum('cgk,bckhw->bgkhw', rq, hp)

def cal_h_attn(q, k):
    # Hadamard product
    hp = q*k
    # (B, G, K*K, H, W)
    return (cal_hp_rk(hp) + cal_rq_hp(hp) + rb.reshape(1, G, K*K, 1, 1)).softmax(2)
    
\end{lstlisting}
\caption{{Strictly follow Equation~\ref{eq:hadamard} (PyTorch-like)}}
\label{alg:strict}
\end{algorithm}

\begin{algorithm}[t]
\definecolor{codeblue}{rgb}{0.25,0.5,0.25}
\lstset{
	backgroundcolor=\color{white},
	basicstyle=\fontsize{7.2pt}{7.2pt}\ttfamily\selectfont,
	columns=fullflexible,
	breaklines=true,
	captionpos=b,
	commentstyle=\fontsize{7.2pt}{7.2pt}\color{codeblue},
	keywordstyle=\fontsize{7.2pt}{7.2pt},
}
\begin{lstlisting}[language=python]
# B: batch size, C: channel size
# H, W: the height / width of feature map
# G: the number of heads, K: kernel size
# q, k: queries and keys in shape (B, C, H, W)

def init()
    # einsum is equivalent to 1x1 convolution
    # rb is equivalent to bias in convolution
    cal_hp_rk = nn.Conv2d(C, G*K*K, 1, bias=True)
    cal_rq_hp = nn.Conv2d(C, G*K*K, 1, bias=True)
    
def cal_h_attn(q, k):
    # Hadamard product
    hp = q*k
    
    hp_rk = cal_hp_rk(hp)
    rq_hp = cal_rq_hp(hp)
    
    kernel = torch.zeros(G*K*K, 1, K, K)
    for i in range(G*K*K):
        _id = i % K*K
        _x = _id % K
        _y = _id // K
        kernel[i, 0, _y, _x] = 1
    rq_hp = F.conv2d(rq_hp, kernel, padding=K//2, groups=G*K*K)
    rq_hp = rq_hp
    
    h_attn = (hp_rk + rq_hp).reshape(B, G, K*K, H, W)
    return h_attn.softmax(2)
    
\end{lstlisting}
\caption{{Equivalent 1 of Equation~\ref{eq:hadamard} (PyTorch-like)}}
\label{alg:equivalent1}
\end{algorithm}

\begin{algorithm}[t]
\definecolor{codeblue}{rgb}{0.25,0.5,0.25}
\lstset{
	backgroundcolor=\color{white},
	basicstyle=\fontsize{7.2pt}{7.2pt}\ttfamily\selectfont,
	columns=fullflexible,
	breaklines=true,
	captionpos=b,
	commentstyle=\fontsize{7.2pt}{7.2pt}\color{codeblue},
	keywordstyle=\fontsize{7.2pt}{7.2pt},
}
\begin{lstlisting}[language=python]
# B: batch size, C: channel size
# H, W: the height / width of feature map
# G: the number of heads, K: kernel size
# q, k: queries and keys in shape (B, C, H, W)

def init()
    # merge lyaers
    conv_1x1 = nn.Conv2d(C, 2*G*K*K, 1, bias=True)
    
def cal_h_attn(q, k):
    # Hadamard product
    hp = q*k

    hp_attn = conv_1x1(hp)
    
    kernel = torch.zeros(G*K*K, 2, K, K)
    for i in range(G*K*K):
        _id = i % K*K
        _x = _id % K
        _y = _id // K
        kernel[i, 0, K//2, K//2] = 1
        kernel[i, 1, _y, _x] = 1
    hp_attn = F.conv2d(hp_attn, kernel, padding=K//2, groups=G*K*K)
    return hp_attn.reshape(B, G, K*K, H, W).softmax(2)
    
\end{lstlisting}
\caption{{Equivalent 2 of Equation~\ref{eq:hadamard} (PyTorch-like)}}
\label{alg:equivalent2}
\end{algorithm}

\end{document}